\documentclass[conference]{IEEEtran}
\IEEEoverridecommandlockouts
\usepackage{cite}
\usepackage{amsmath,amssymb,amsfonts}
\usepackage{algorithmic}
\usepackage{graphicx}
\usepackage{textcomp}
\usepackage{xcolor}
\usepackage[colorlinks,urlcolor=blue,linkcolor=blue,citecolor=blue]{hyperref}
\def\BibTeX{{\rm B\kern-.05em{\sc i\kern-.025em b}\kern-.08em
    T\kern-.1667em\lower.7ex\hbox{E}\kern-.125emX}}
\begin{document}

\title{ Deep Learning-based Spatio Temporal Facial Feature Visual Speech Recognition
 \\
}

\author{\IEEEauthorblockN{
\large Pangoth Santhosh Kumar \IEEEauthorrefmark{1}, Garika Akshay\IEEEauthorrefmark{1}\\
\IEEEauthorblockA{\IEEEauthorrefmark{1}\normalsize Department of Data Science and Artificial Intelligence, IIIT Naya Raipur, India}
\normalsize {e-mail:  pangoth20102@iiitnr.edu.in, garika20102@iiitnr.edu.in}
}}

\maketitle

\begin{abstract}
In low-resource computing contexts, such as smartphones and other tiny devices, Both deep learning and machine learning are being used in a lot of identification systems. as authentication techniques. The transparent, contactless, and non-invasive nature of these face recognition technologies driven by AI has led to their meteoric rise in popularity in recent years. While they are mostly successful, there are still methods to get inside without permission by utilising things like pictures, masks, glasses, etc. In this research, we present an alternate authentication process that makes use of both facial recognition and the individual's distinctive temporal facial feature motions while they speak a password. Because the suggested methodology allows for a password to be specified in any language, it is not limited by language. The suggested model attained an accuracy of 96.1\% when tested on the industry-standard MIRACL-VC1 dataset, demonstrating its efficacy as a reliable and powerful solution. In addition to being data-efficient, the suggested technique shows promising outcomes with as little as 10 positive video examples for training the model. The effectiveness of the network's training is further proved via comparisons with other combined facial recognition and lip reading models.

\end{abstract}


\section{Introduction}
Digital systems have long used biometric authentication technologies to identify authorised users and provide them access to protected resources. Verifying a user's identification using a photograph of their face has been more popular in recent years due to the system's contact-less\cite{article1}, noninvasive nature and low barrier to entry. Hand-engineered features like SIFT, LBP, and Fisher vectors formed the foundation of early image-based face recognition systems. Deep learning approaches like FaceNet , Baidu \cite{DBLP:journals/corr/abs-1902-03524}, and DeepID models \cite{DBLP:journals/corr/abs-2012-02515} have recently eclipsed human performance in terms of accuracy. Despite this, unauthorised individuals have often been able to trick face authentication systems by utilising a picture of a legitimate user. To address this weakness, researchers are looking at models that analyse the user's lip movements as they say a word. LipNet \cite{gergen16_interspeech}, a deep learning network for lip reading, has reached 95\% accuracy in sentence-level categorization.
Lip reading models based on HMM, LSTM Networks, CNNs, and other neural network architectures have also been developed. However, they are too dependent on lip movement and lip characteristics and struggle to adapt to varied lighting situations\cite{singh2022video}.

\begin{figure*}[ht]
    \centering
    \centerline{\includegraphics[width=\linewidth]{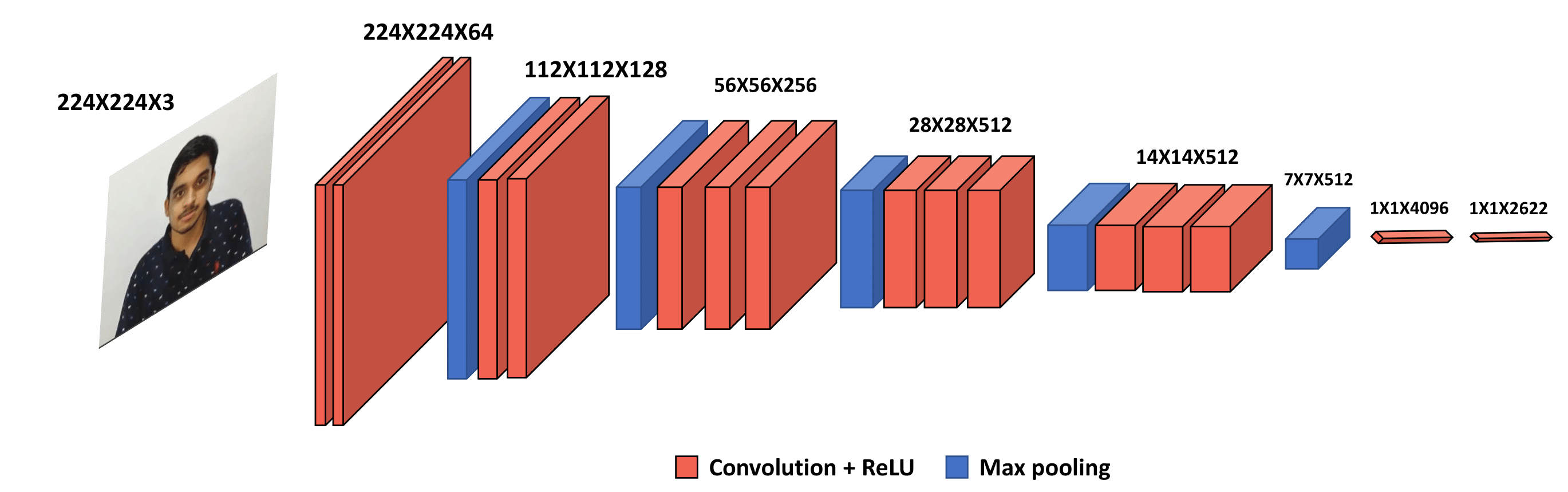}}
    \caption{Proposed Architecture built on VGGFace}
    \label{LSTM}
\end{figure*} 

In order to solve the security and privacy issues, a novel solution has been developed that combines deep facial recognition models with photos without the requirement for audio. In this research, we present an authentication model that records the dynamic patterns of a person's face when they say a certain phrase. The facial characteristics were captured using a VGGNet model \cite{liu2015targeting} dubbed VGGFace \cite{7486599} that had been pre-trained on faces; the sequence of features was then processed through numerous LSTM layers, and the model learned to predict whether a genuine user was saying their chosen password or not. The selected password is not given to the model, just videos of it being said. This means that if pronounced continuously, it may be any alphabetic or numeric sequence. To evaluate the efficacy of the proposed model on real-world data, it was benchmarked using the publicly accessible standard dataset MIRACL-VC1 \cite{937709}.

Tend to get around security measures by utilising a picture of the real user. To address this weakness, researchers are looking at models that analyse the user's lip movements as they say a word. Sentence-level classification accuracy of 95\% has been attained by deep learning models for lip reading like LipNet \cite{inproceedings}.
Lip reading models based on HMM, LSTM Networks, CNNs, and other neural network architectures have also been developed. However, they are too dependent on lip movement and lip characteristics and struggle to adapt to varied lighting situations.
In order to solve the security and privacy issues, a novel solution has been developed that combines deep facial recognition models with photos without the requirement for audio. In this research, we present an authentication model that records the dynamic patterns of a person's face when they say a certain phrase. To do this, we first used a VGGNet model \cite{ouyang2015deepidnet} dubbed VGGFace \cite{Parkhi15} have been trained on expressions to learn the features of faces, afterwards we fed it pictures of people, the sequence of features through numerous LSTM layers, and the model learned to predict whether or not a genuine user was reciting their chosen password. The selected password is not given to the model, just videos of it being said. This means that if pronounced continuously, it may be any alphabetic or numeric sequence. To evaluate how well the proposed model performs on real-world data and requirements, a second dataset was built, consisting of videos captured by smartphone cameras in a variety of environments. This dataset was then benchmarked against the publicly available standard dataset, MIRACL-VC1 \cite{DBLP:journals/corr/abs-1902-03524}. This compiled dataset was also used to test the system.

The fundamental contribution of this work is a secure video-based face authentication system that utilises a video of a person speaking a password to detect and prevent imposter scenarios, such as i) the same person pronouncing an alternative phrase (Same Person scenario) or, ii) an alternative individual attempting to obtain permission
 ( Different person case).This model's independence from a particular language and domain is another important addition and feature. As far as we're aware, no such system has ever been put into place.
This paper will continue in the following format. In Section 2, we'll go through the recent research that's been done on lip reading and authentication, as well as the current limitations that call for new developments. The methodology presented is described in Section 3, followed by an evaluation and validation of the model in Section 4, a summary and a list of references in Section 5.

\section{Related works}
Several widely used models for facial recognition based on deep learning have been presented recently\cite{DBLP:journals/corr/abs-2012-02515}.
Near-perfect results have been achieved on benchmark datasets by the FaceNet \cite{7298682} and DeepID models \cite{DBLP:journals/corr/SunWT14} \cite{inproceedings}. The performance of these facial recognition systems, which have been trained on millions of photos, has exceeded that of humans. When employed as authentication systems, however, they remain vulnerable to assaults. This section presents the current state of the art efforts in this area and identifies the few holes that have been found in the performance of these systems.

\cite{inproceedings} investigated lip reading methods for word categorization and lip password authentication. Traditional algorithms like FCM and LCACM are employed for lip localization in a picture, however they aren't the only ones that have been tried.
They found that at the time, Gaussian Mixture Models (GMMs) and Hidden Markov Models were the most popular classifiers for visual speech detection.
(HMMs). Based only on the behavioural biometrics of lip movements,\cite{liu2015targeting} suggested a lip password verification model using Multi-boosted HMMs. They used an algorithm that divides lips to retrieve visual sequences of the mouth area, then used Multiboosted HMMs on those subunits to determine a decision boundary. However, users were limited to passwords consisting entirely of digits, resulting in an equal error rate (EER) of 4.06\%. According to the work of \cite{ouyang2015deepidnet}, a CNN trained on pictures of the mouth can be used to forecast phenomes in visual speech recognition systems based on deep learning. For the purpose of the word recognition task, the CNN outputs were treated as sequences, and an HMM+GMM observation model was used. Their results showed that visual characteristics learned using a CNN generalised across domains performed better than more standard methods.

suggest LSTM-based lip reading. \cite{DBLP:journals/corr/ChungSVZ16}. Using a Histogram of Oriented Gradients, the LSTM classified words from sequences of mouth photos. (HOG). They outperformed SVM and HMM classifiers. \cite{Amit2016LipRU} built a hybrid CNN-LSTM model for word classification on the same MIRACL-VC1 dataset as this work. CNN+LSTM, CNN, and SVM classifiers were compared. The CNN method performed best for visual speech recognition (61.2\%), while the CNN+LSTM model may improve with extra training.
LipNet, created by \cite{shillingford2018largescale}, accurately categorises sentence-level lip reading. Spatio-temporal convolutions \cite{medi2023novel} and recurrent neural networks helped them outperform human lip readers.
In the recent times, vision transformers-based architectures \cite{krishna2022vision}\cite{krishna2023lesionaid} have been performing better than neural network architectures in computer vision. This paper points out the performance of vision transformers on Lip reading.

Through discussion on many past research and state-of-the-art models, several shortcomings in facial recognition-based biometric systems were uncovered. Like any technology, face recognition systems may be exploited and have security weaknesses. Lip password models are inaccurate, and passwords must be solely numbers. Most models only speak their training language. This paper develops a language- and domain-neutral methodology to alleviate these limits. Next, we'll explain the model.

\section{Research Methodology}
Our proposed model is similar to AuthNet\cite{raghavendra2020authnet} and employs the two-stage convolutional neural network architecture shown in Fig.\ref{Proposed archi}. At each frame, we extract key face traits using a VGGFace model that has already been trained \cite{Parkhi15}. A tiny network of stacked LSTM layers is used to recognise and learn the behaviours of these characteristics. After training, this network can tell whether the person-word combination in a test case is the same as the one that was trained for.

There are 10 speakers in the MIRACL-VC1 Words dataset, and they each say 10 words 10 times. This is broken down into sub-sections for ease of testing and validation. To test the model's resilience against the Different Person Case, 5 speakers, or 5 × 10 x 10 = 500 utterances, were arbitrarily put aside as unseen data. (mentioned above). To evaluate the model's performance in the Same Person Case, we put aside 3 words from each of the remaining 5 speakers (5 x 3 x 10 = 150 utterances). As a result, 650 specimens were removed from consideration. Five times seven times ten is 350 remaining utterances, 10 of which are good (right speaker, correct word), 60 are negative (right speaker, wrong word), and 280 are wrong speakers. Oversampling the 10 positive samples results in 100 more positive samples, for a grand total of 350 + 100 = 450 instances. This is then divided into a train and test set with a 70:30 split; the 650 instances that were previously reserved are added to the test set as negative samples to account for the aforementioned impostor scenarios.This method is used to guarantee that we are working with information that the model has never seen before.
Ten samples of correct person and correct word (oversampled to 110), ninety samples of correct person saying wrong word, ninety samples of wrong person saying correct word, and eight hundred and ten samples of wrong person saying wrong word add up to a total of one thousand words (10 speakers uttering 10 words 10 times each).

The model was trained in an iterative way on all such conceivable combinations and evaluated against unseen cases to offer a full analysis of the performance and show the generalizability and resilience of the system. Therefore, there are 35 potential person-word combinations generated by repeating the training process for the 5 randomly selected speakers who each pronounce 7 unique words.

\subsection{Data preprocessing.}\label{AA}
\begin{figure}[ht]
    \centering
    \includegraphics[width=\linewidth]{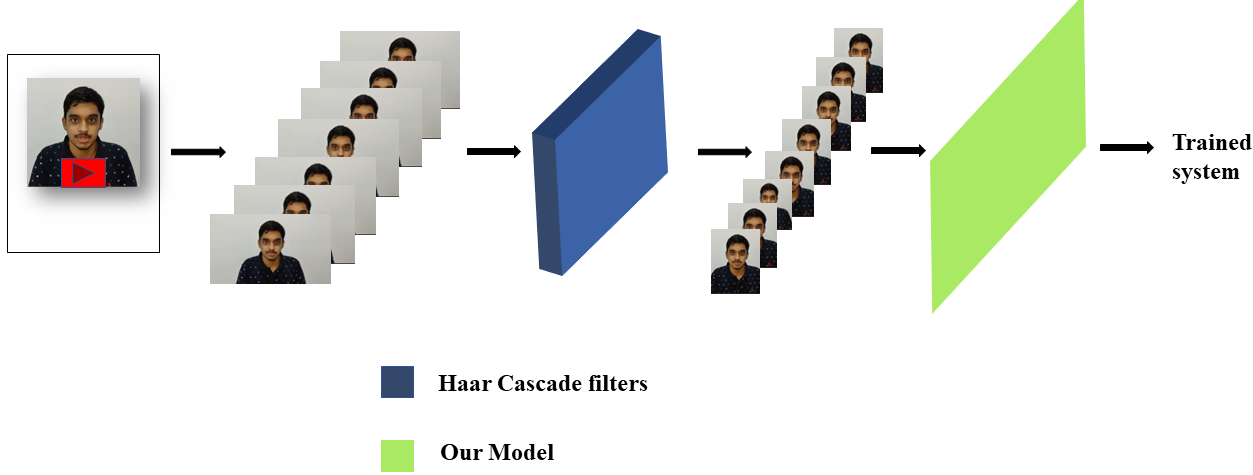}
    \caption{Data Preprocessing Pipeline }
    \label{pipe}
\end{figure}

In the MIRACL-VC1 dataset, each testing and training sample is made up of 5-15 photographs of the individual saying a password (see Fig.\ref{pipe}). Only the words' colour images have been used for training so far\cite{raghavendra2020authnet}. Pictures of people are first sent through a Haar cascade face detector \cite{937709} before they are cropped. The Haar cascade detector, which is part of OpenCV, is often used to identify people and their various facial features. You may use it for basic item identification as well as recognising faces. The four primary phases of the strategy are selecting Haar features, creating a collection of images for easier being processed, training using Adaboost, and lastly feeding the data through a sequence of cascade classifiers. Padding images with white photographs ensures that all frames have the same number of frames. In order for the trained VGGNet model to handle the photos, they are downscaled to a size of 224 x 224. Identical procedures were used for the manual creation of the dataset. This set of numbers represents a large portion of one possible multiverse. Before being fed via the VGGFace framework, the frames are separated at an even frame rate, padded to keep the number of timesteps constant, and scaled to a resolution of 224 by 224.

\subsection{VGGFace}
A vast series of convolutional layers trained on hundreds of thousands of photos of celebrity faces makes up the pre-trained VGGFace model\cite{9940584}, which is applied to each image. Each image's feature vector has an output of 2,622 dimensions. To prepare the data for input into the LSTM layer, this procedure is repeated for each individual and each phrase (as shown in Fig. \ref{LSTM}). Each sample therefore consists of 2622 characteristics over 20 timesteps. This is a binary classification issue since the data are marked with a 1 for the right personword combination and a 0 for all others. Oversampling was used to address the ensuing inequity in the representation of different social groups.

\subsection{LSTM network}
When dealing with sequential data, LSTM networks have shown remarkable success in a broad range of application fields. To do this, we use a 4 LSTM layer network where each layer has 20-time steps in order and an output sigmoid layer to provide probabilistic forecasts. After being trained, this model may decide whether or not a given test video has the required person-word combination before granting access to a verified user.

\section{EXPERIMENT RESULTS}
In order to demonstrate that the proposed strategy is effective, a number of tests were carried out. In order to train the model, an i7-8640U central processing unit and 8 gigabytes of RAM were used. Both the conventional MIRACL-VC1 dataset and a constructed dataset of videos, each of which is 2 seconds in duration, were used in the evaluation of the model. The Realme X3 superzoom (4k, 64MP), the OnePlus Nord (4k, 32MP), and the Apple iPhone 13 were utilised to record the clip on their respective devices. (2k, 16MP). They were shot in a variety of locations with variable lighting and backdrops. Table.\ref{Data Description} summarises each dataset's key statistics.
The average time for the suggested model to process a person-word pair was 154.2 seconds.
\begin{figure*}[ht]
    \centering
    \centerline{\includegraphics[width=\linewidth]{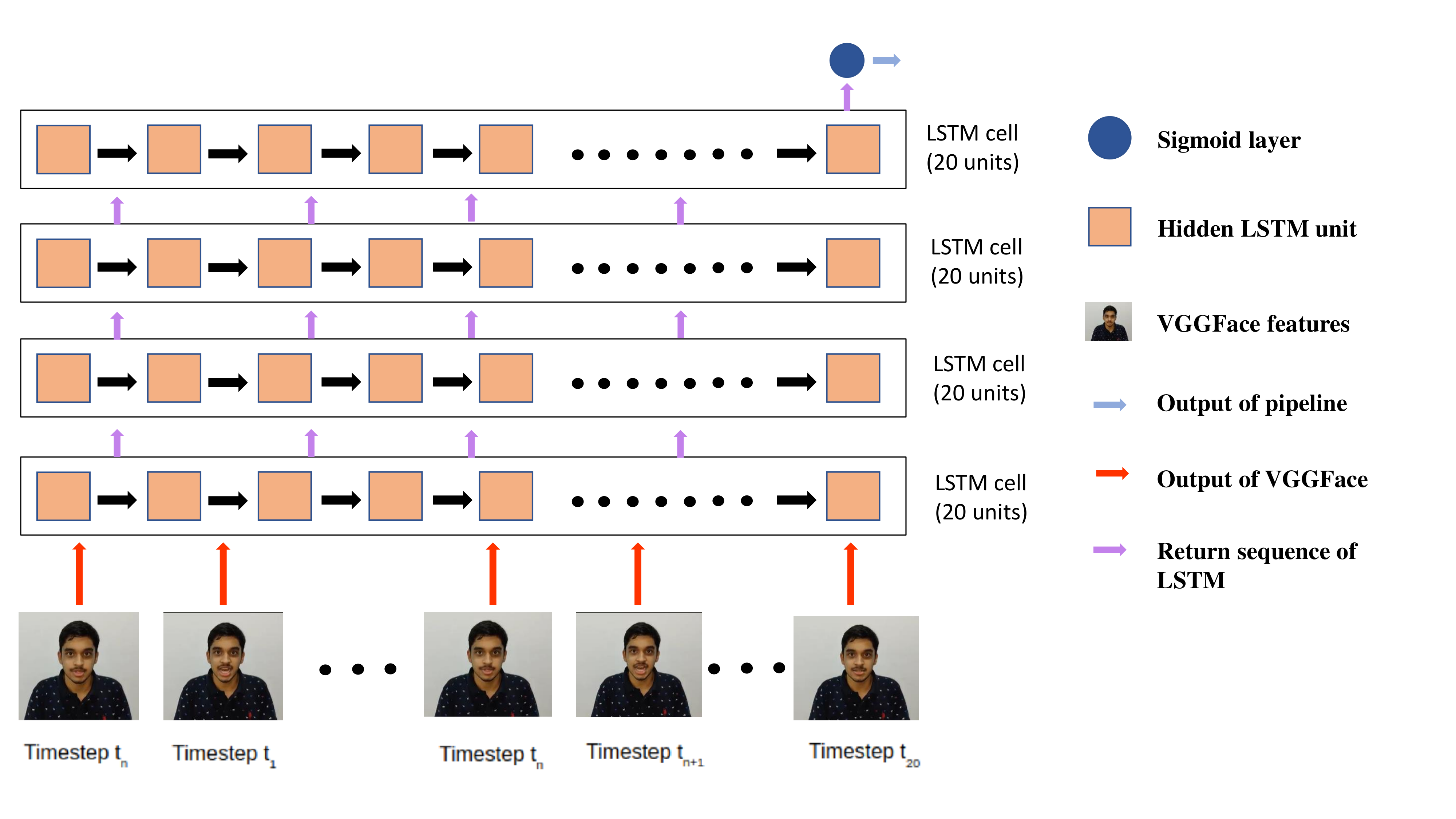}}
    \caption{Proposed Architecture built on LSTM Network}
    \label{Proposed archi}
\end{figure*} 

\begin{table}[ht]
    \centering
        \caption{data description}
\begin{tabular}{|l|c|c|}
\hline Dataset & MIRACL-VC1 & Data Set Assembled \\
\hline i. Words & 10 & 3 \\
\hline ii. Speakers & 10 & 3 \\
\hline iii. Utterances & 10 & 6 \\
\hline iv. Image Resolution & $640 \times 480 \times 3$ & $350 \times 640 \times 3$ \\
\hline v. Language used & Engl & Engl and Hin \\
\hline
\end{tabular}
    \label{Data Description}
\end{table}

The purpose of the suggested pipeline is to perform a binary classification job; the pipeline is also intended to accept as input a sequence of pictures with a resolution of 640 by 480 pixels, which are arranged in the order of the timesteps. After that, the photos were processed using the pretrained version of the VGGFace model, as shown in Fig.\ref{pipe}. Following this, the generated feature vectors were input into the LSTM network for the purposes of training and assessing it, as shown in Fig.\ref{Proposed archi}. The proposed model was trained utilising the binary crossentropy (BCE) loss function (derived as per Eq.(\ref{bce})) and the Adam optimizer, with an initial learning rate of 0.001 over the course of 60 epochs and a batch size of 75\cite{raghavendra2020authnet}.

\begin{equation}
\mathbf{BCE}=-\frac{1}{n} \sum_{i=1}^n\left(y_i \cdot \log \left(\hat{y}_i\right)+\left(1-y_i\right) \cdot \log \left(1-\hat{y}_i\right)\right)
\label{bce}
\end{equation}

\section{Performance Metrics}
Many different standard measures were utilised to assess the performance of the proposed model and prove its effectiveness. The authorization ratio of the framework, or the probability that an authenticated user will be permitted access, is based on the sensitiveness of the framework, which may be calculated using Eq.(\ref{re}) The rejection ratio of a system is its specificity or the probability that an unwanted attacker would be denied entry to the system (as shown by the Eq.(\ref{spe})). It represents the overall efficacy of the authentication system and is used to evaluate the model's performance (through Eq. (\ref{acc})). The degree of accuracy achieved is indicative of its effectiveness.

\begin{equation}
\text { Recall }=\frac{T P}{T P+F N} \\
\label{re}
\end{equation}

\begin{equation}
\text { Specificity }=\frac{T N}{T N+F P} \\
\label{spe}
\end{equation}

\begin{equation}
\text { Accuracy }=\frac{T P+T N}{T P+T N+F P+F P}
\label{acc}
\end{equation}

\begin{figure}[ht]
    \centering
    \includegraphics[width=\linewidth]{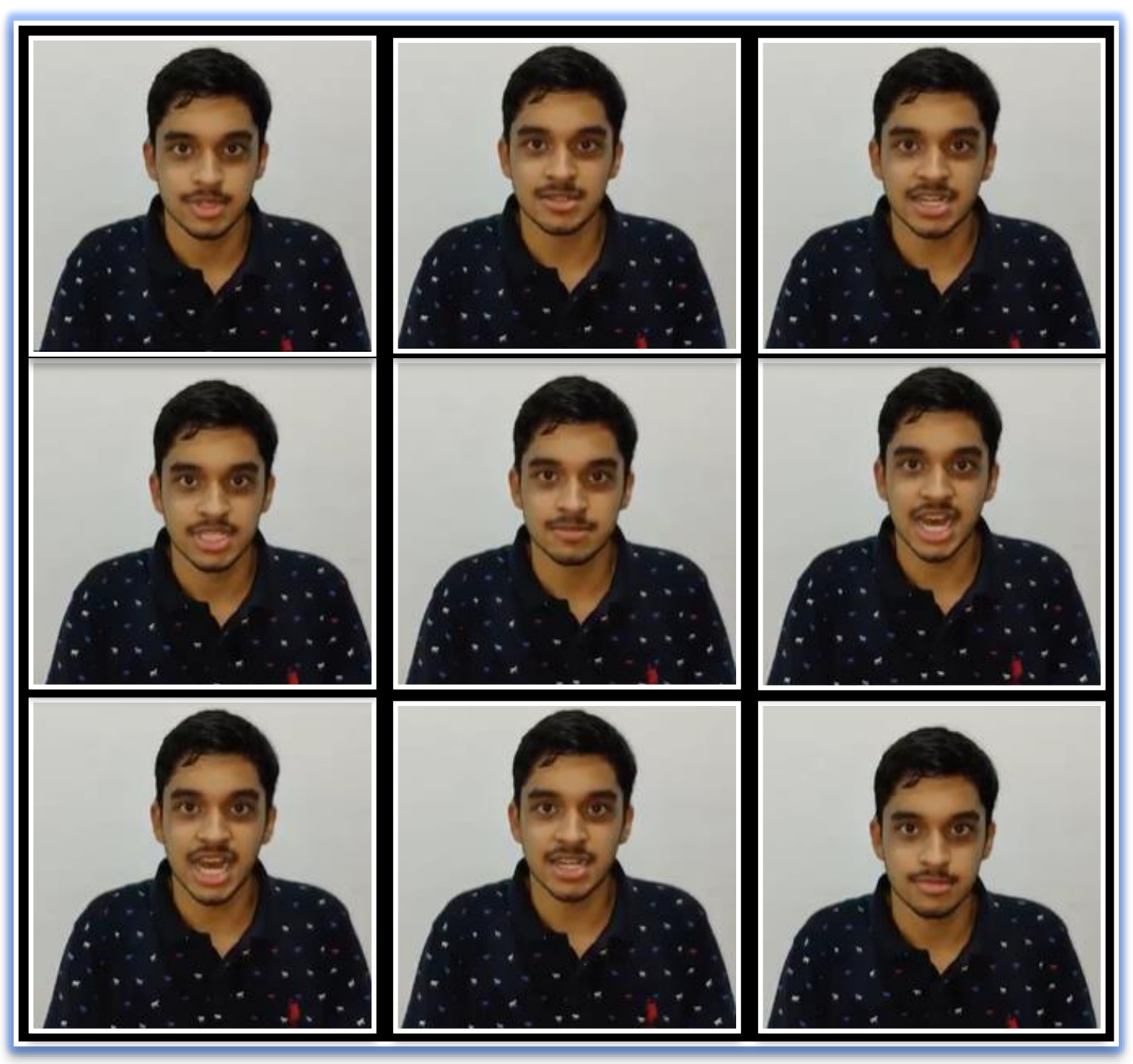}
    \caption{Variations in future extraction}
    \label{NI}
\end{figure}

True Positive and True Negative samples are indicated by the acronyms TP and TN, respectively. On the other hand, False Positive and False Negative samples are indicated by the abbreviations FP and FN, respectively. During the process of examining the many different components of the model that was provided, two more metrics were used. The Equal Error Rate (EER) metric is stated to have been attained at the point in time when the probability of an unauthorised user being refused access is equal to the chance of an intruder acquiring entrance. This is the point at which the likelihood of an unauthorised user being denied access is equal to the probability of an intruder obtaining admittance. It is also the point at which the False Acceptance Ratio (FAR) is comparable to the False Rejection Rate. This is because of the reason stated above. (FRR). Calculating the FAR is done using Eq.(\ref{far}), while computing the FRR is done using Eq.(\ref{frr}). The Equivalence Entropy Ratio (EER) is a standardised measure that is used for the purpose of comparison between different biometric systems. The Receiver Operating Characteristic curve, often known as the ROC, was used in order to determine the Effective Exposure Rate (EER) since it is less susceptible to scaling modifications. The ROC Curve has been plotted, and Fig.\ref{roc} provides a depiction of what it looks like. 

\begin{equation}
\begin{aligned}
\mathbf{F A R} & =\frac{F P}{T P+T N+F P+F N} \\
\end{aligned}
\label{far}
\end{equation}

\begin{equation}
\begin{aligned}
\mathbf{F R R} & =\frac{F N}{T P+T N+F P+F N}
\end{aligned}
\label{frr}
\end{equation}

The ROC curve shows how the True Positive Rate (TPR) and the False Positive Rate (FPR or FAR) alter in relation to one another when the threshold is varied. The receiver operating characteristic (ROC) curve may be seen as a cost-benefit analysis when making decisions. To ensure the pipeline gets the most true positives possible for each false positive it encounters, it's also utilised to compute the threshold. This ensures that the greatest number of legitimate accesses are provided by the system while minimising the possibility of any unauthorised access being granted. Finding the intersection of the ROC curve with the line (x, y) = 1 gives us the point of equal error. The Receiver Operating Characteristic (ROC)

\cite{raghavendra2020authnet}curve is used in the Area Under the ROC curve computation. (AUC). The AUC measures how likely it is that the model would assign a higher score to a hypothetically positive example than to a hypothetically negative one. Area under the curve (AUC) is a metric that measures how confident a model is in its decision boundaries and does not need a threshold.

\section{Results and Analysis}
To measure our proposed methodology efficacy, researchers built a cross-validation dataset that included three types of imposter cases for any given person-word pair: the same speaker using a different word, a different speaker using the same word, and a third speaker using a different word. The voices used in the test were selected at random from a pool of people the model has never seen before throughout the training process. 

\begin{figure*}[ht]
    \centering
    \centerline{\includegraphics[width=\linewidth]{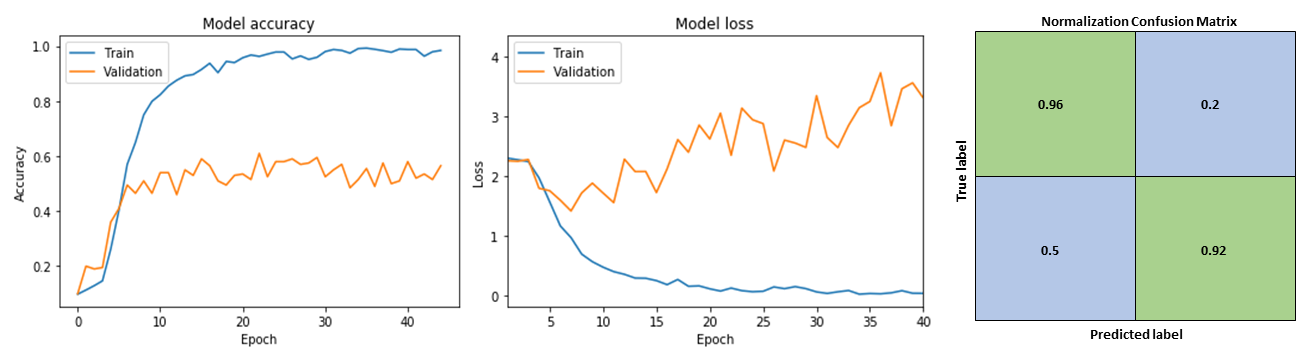}}
    \caption{Learning Curves of Training and Validation Accuracy and Loss for 30 Epochs and its Confusion Matrix}
    \label{learning}
\end{figure*}

The model's capacity to generalise to new person-word pairings is also rigorously evaluated by selecting a set of terms that are wholly novel yet are all uttered by the same individual. Table.\ref{performance} compares the proposed model's performance to that of other metrication strategies, and Fig.\ref{learning} displays the associated confusion matrix.

\begin{equation}
\cos (\theta)=\frac{\mathbf{A} \cdot \mathbf{B}}{\|\mathbf{A}\|\|\mathbf{B}\|}=\frac{\sum_{i=1}^n A_i B_i}{\sqrt{\sum_{i=1}^n A_i^2} \sqrt{\sum_{i=1}^n B_i^2}}
\label{cos}
\end{equation}

The differences between the photographs were calculated using the Cosine Difference, which was calculated using Eq. (\ref{cos}). The cosine difference between feature vectors serves. The purpose of comparing feature vectors acquired by the VGGFace model across speakers and words, or even across time for the same speaker, is to demonstrate that there is a substantial variation between the feature vectors.
The empirical rationale for how facial characteristics vary when people recite their passwords is provided by this variation in feature vectors between individuals and words. Therefore, the LSTM network is able to capture substantial variations across photos when the pre-trained VGGFace model extracts a feature vector for each image. This is seen in Fig.\ref{NI}, and it emphasises how the suggested model can deal with imposters saying the same password, enabling it to function as an open password system.

\begin{figure}[ht]
    \centering
    \includegraphics[width=\linewidth]{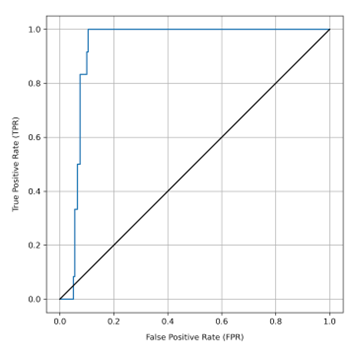}
    \caption{ROC Curve for the MIRACL-V1 dataset}
    \label{roc}
\end{figure}

The system's high sensitivity of 0.96, as shown in Table.\ref{performance}, indicates that it incorrectly identifies an input user-password combination just 2 times for every 1000 attempts. It's an excellent gauge of our ability to distinguish false positives. The model's high specificity shows how well it can identify different types of imposter attacks. The varying degrees of wrong specificity for each imposter circumstance are shown in Table.\ref{error}. The data shows that although the system is successful in preventing attacks from a new user almost all the time (about 96 percent of the time), it is only able to correctly recognise an erroneous password being provided around 9 out of every 10 times. This proves the model can serve as a robust defensive mechanism against attacks from malicious actors, ensuring a high level of personal privacy is preserved. Fig.\ref{roc} shows the ROC curve, from which we can calculate that the EER is 0.023 by selecting the point at which the TPR is 0.963 and the FPR is 0.023. The equal error rate of a pipeline, which should be reasonably low, may be used as a proxy for its ability to reduce the amount of false acceptances and incorrect rejections. The pipeline seems to be rather confident in its prediction, with an area under the curve (AUC) of 0.96.

\begin{table}[]
        \caption{Comparision of proposed and existing works}
\centering
\begin{tabular}{|c|c|}
\hline \text { Existing } & \text { Test Accuracy } \\
\hline \text { (amit, jnoyola, and sameepb 2016) } & 0.631 \\
\hline \text { (Chung et al. 2017) } & 0.682 \\
\hline \text { (Gergen et al. 2016) } & 0.853 \\
\hline \text { (Wand, Koutnik, and Schmidhuber 2016) } & 0.945 \\
\hline \text { ViViT on MIRACL-VC1 \cite{singh2022video} } & 0.945 \\
\hline \text { Proposed } & 0.960\\
\hline

\end{tabular}
\label{comparision}
\end{table}

We find that the model performs similarly on the MIRACL-VC1 dataset developed in the lab as it does on the collected dataset acquired from videos shot by a smartphone camera. This displays the system's flexibility in dealing with pictures and lighting conditions seen in the actual world. This suggests that the system may inadvertently pick up patterns that are unique to each user, and that the results are not just the result of overfitting to data obtained in a lab.

\begin{table}[]
        \caption{performance of proposed methodology}
\centering
\begin{tabular}{|c|c|c|}
\hline\text { Metric } & \text { MIRACL-VC1 } & \text { Collated Dataset } \\
\hline \text { Sensitivity } & 0.988 & 0.977 \\
\hline\text { Specificity } & 0.959 & 0.966 \\
\hline\text { Accuracy } & 0.960 & 0.955 \\
\hline\text { AUC Score } & 0.990 & 0.989 \\
\hline\text { Equal Error Rate } & 0.021 & 0.030 \\
\hline
\end{tabular}

    \label{performance}
\end{table}

As evidence of the efficiency of the training, we compare our model to a two-tiered system consisting of various state-of-the-art models in face recognition and lip reading Fig.\ref{Proposed archi}. 

The state-of-the-art FaceNet model \cite{7298682} was used as a benchmark since it claims an accuracy of 99.76 percent. A lip reading model was presented by \cite{Amit2016LipRU} with an accuracy of 61.2\%. \cite{DBLP:journals/corr/ChungSVZ16} and \cite{DBLP:journals/corr/SunWT14} models, which achieved 65.4\% and 86.6\% accuracy, respectively, 
were the previous state-of-the-art models in lip reading on the word level. The suggested model's efficacy has been measured against that of Lipnet \cite{shillingford2018largescale}.

\begin{table}[ht]
    \centering
        \caption{Error classification granularity (MIRACLVC1 dataset)}

\begin{tabular}{|c|c|c|}

\hline \text {Varieties of Forgery/Cases of Error} & \text { Data Samples } & \text { Specificity } \\
\hline \text { Identical Speaker-Varying Tone } & 200 & 0.9265 \\
\hline\text { Similar Expressions-Varying Speaker } & 135 & 0.9687 \\
\hline\text { Distinct Individuals-Distinct Expressions } & 90 & 0.9440 \\
\hline
\end{tabular}

    \label{error}
\end{table}

Table.\ref{comparision} displays the results and comparisons, demonstrating that the trained model achieves the same performance as the sum of state-of-the-art models while protecting against any imposter assaults or faults present in the latter.

\section{Conclusion and Future Work}
To verify people using their facial movements over time, we developed a deep neural network that combines a convolutional neural network (CNN) and a recurrent neural network (RNN). Our model was tested on the MIRACL-VC1 dataset, and it was shown to be 96.1\% accurate there. Cross-validation was used to verify the dataset's usefulness and demonstrate that it is not language-specific. We also showed that our system can be tailored to individual tastes in terms of background visuals, lighting, and video quality, making it particularly well-suited to mobile and low-resource smart devices. Our methodology might be used as an open credential system in which the danger of disclosing the password is eliminated if it can accurately recognise the same password being uttered by different persons. Since our method only needs temporal facial movements captured from speech videos and not information about the speaker's dialect, it is not hindered by language barriers. Our method showed exceptional efficiency in terms of data needs, requiring just 100 good video examples for training. Currently, each sample takes about 10 seconds to test, but with some optimisation, that time could be cut in half. The development of an authentication system for mobile devices and personal computers presents a substantial difficulty in optimising testing time.

\bibliographystyle{IEEEtran}
\bibliography{main}

\end{document}